\pdfoutput=1
\documentclass[runningheads]{llncs}
\usepackage{graphicx}

\usepackage{tikz}
\usepackage{comment}
\usepackage{amsmath,amssymb} 

\usepackage{mathtools}
\usepackage[pagebackref,breaklinks,colorlinks,bookmarks=false]{hyperref}
\usepackage{breakcites}
\usepackage{booktabs}

\usepackage{xcolor}

\usepackage[accsupp]{axessibility}  
\usepackage{array}
\newcolumntype{H}{>{\setbox0=\hbox\bgroup}c<{\egroup}@{}}

\usepackage{multicol}
\usepackage{multirow}
\usepackage{subfig}
\usepackage{wrapfig}

\newcommand{\tempmeth}{HTC }
\newcommand{\ensmeth}{SCM }

\newcommand{\multiplier}{m}

\begin{document}
\pagestyle{headings}
\mainmatter
\def\ECCVSubNumber{7874}  

\title{Distilling the Undistillable: Learning from a Nasty Teacher} 
\makeatletter
\newcommand{\ssymbol}[1]{\@fnsymbol{#1}}
\makeatother

\titlerunning{Distilling the Undistillable}
%
\titlerunning{Distilling the Undistillable}
%
\author{Surgan Jandial\inst{1}\and
Yash Khasbage\inst{2}\and
Arghya Pal\inst{3}\and 
Vineeth N Balasubramanian \inst{2} \and
Balaji Krishnamurthy \inst{1}
}
%
\authorrunning{S. Jandial et al.}
%
\institute{Indian Institute of Technology, Hyderabad \and
Adobe MDSR Labs \and
Harvard University
}
%
\authorrunning{S. Jandial et al.}
%
\institute{Adobe MDSR Labs \\ \and  Indian Institute of Technology, Hyderabad \\ \and 
Dept. Of Psychiatry and Radiology, Harvard \\
}
\maketitle

\begin{abstract}
The inadvertent stealing of private/sensitive information using Knowledge Distillation (KD) has been getting significant attention recently and has guided subsequent defense efforts considering its critical nature. Recent work \textit{Nasty Teacher} proposed to develop teachers which can not be distilled or imitated by models attacking it. However, the promise of confidentiality offered by a nasty teacher is not well studied, and as a further step to strengthen against such loopholes, we attempt to bypass its defense and steal (or extract) information in its presence successfully. Specifically, we analyze \textit{Nasty Teacher} from two different directions and subsequently leverage them carefully to develop simple yet efficient methodologies, named as \textit{HTC} and \textit{SCM}, which increase the learning from Nasty Teacher by upto 68.63\% on standard datasets. Additionally, we also explore an improvised defense method based on our insights of stealing. Our detailed set of experiments and ablations on diverse models/settings demonstrate the efficacy of our approach. 

\end{abstract}

\section{Introduction}
\label{sec:intro}
Knowledge Distillation utilizes the outputs of a pre-trained model (i.e teacher) to train a generally smaller model (i.e student). Typically, KD methods are used to compress models that are wide, deep and require significant computational resources and pose challenges to model deployment.
Over the years, KD methods have seen success in various settings beyond model compression including few-shot learning \cite{fskd}, continual learning {\cite{delange2021pami-cl-survey}}, and adversarial robustness \cite{ard}, to name a few -- highlighting its importance in training DNN models. However, recently, there has been a growing concern of misusing KD methods as a means to steal the implicit model knowledge of a teacher model that could be proprietary and confidential to an organization.  
KD methods provide an inadvertent pathway for leak of intellectual property that could potentially be a threat for science and society. Surprisingly, the importance of defending against such KD-based stealing was only recently explored in \cite{nasty,skeptical}, making this a timely and important topic. 


In particular, \cite{nasty} recently proposed a defense mechanism to protect such KD-based stealing of intellectual property using a training strategy called the `Nasty Teacher'. This strategy attempts to transform the original teacher into a model that is `\textit{undistillable}', i.e., any student model that attempts to learn from such a teacher gets significantly degraded performance. This method maximally disturbs incorrect class logits (a significant source of model knowledge), which produces confusing outputs devoid of clear, meaningful information. This method showed promising results in defending against such KD-based stealing from DNN models. However, any security-related technology development requires simultaneous progress of both attacks and defenses for sturdy progress of the field, and eventually lead to the development of robust models. In this work, we seek to test the extent of the defense obtained by the `Nasty Teacher' \cite{nasty}, and show that it is possible to recover model knowledge despite this defense using the logit outputs of such a teacher. Subsequently, we leverage the garnered insights and propose a simple yet effective defense strategy, which significantly improves defense against KD-based stealing. 

To this end, 
we ask two key questions: (i) can we transform the outputs of the Nasty Teacher to reduce the extent of confusion, and thus be able to steal despite is defense? and (ii) can we transform the outputs of the Nasty Teacher to recover hidden essential relationships between the class logits? To answer these two questions, we propose two approaches -- High-Temperature Composition (HTC) which systematically reduces confusion in the logits and Sequence of Contrastive Model (SCM) which systematically recovers relationships between the logits. These approaches result in performance improvement of KD, thereby highlighting the continued vulnerability of DNN models to KD-based stealing. Because of their generic formulation and simplicity, we believe our proposed ideas could apply well to similar approaches that may be developed in future along the same lines as the Nasty Teacher. To summarize, this work analyzes key attributes of output scores (which capture the strength and clarity of model knowledge) that could stimulate knowledge stealing and thereby leverages those to strengthen defenses against such attacks too. Our key contributions are summarized as follows:
\begin{itemize}
    \item We draw attention to the recently identified vulnerability of KD methods in model-stealing, and analyze the first defense method in this direction, i.e. Nasty Teacher, from two perspectives: (i) reducing the extent of confusion in the class logit outputs; and (ii) extracting essential relationship information from the class logit outputs. We develop two simple yet effective strategies -- High Temperature Composition (HTC) and Sequence of Contrastive Model (SCM) -- which can undo the defense of the Nasty Teacher, pointing to the need for better defenses in this domain. 
    \item Secondly, we leverage our obtained insights and propose an extension of Nasty Teacher, which outperforms the earlier defense under similar settings.
    \item We conduct exhaustive experiments and ablation studies on standard benchmark datasets and models to demonstrate the effectiveness of our approaches.
\end{itemize}
We hope that our efforts in this work will provide important insights and encourage further investigation on a critical problem with DNN models in contemporary times where privacy and confidentiality are increasingly valued.
\section{Related Work}
\label{sec:related-work}
We discuss prior work both from perspectives of Knowledge Distillation (KD) as well as its use in model-stealing below.

\noindent \textbf{Knowledge Distillation:} KD methods transfer knowledge from a larger network (referred to as \textit{teacher}) to a smaller network (referred to as \textit{student}) by enforcing students to match the teacher's output. With seminal works \cite{bucilua,hinton} laying the foundation, KD has gained wide popularity in recent years. The initial techniques for KD mainly focused on distilling knowledge from logits or probabilities. This idea got further extended to distilling features in \cite{fitnet,at,fsp,rkd}, and many others. In all such methods, KD is used to improve the performance of the student model in various settings. More detailed surveys on KD can be found in \cite{survey1, survey2, survey3}. Our focus in this work, however, is on recent works {\cite{nasty,dream2distill,skeptical}}, which have discussed how KD can unintentionally expose threats to Intellectual Property (IP) and private content of the underlying DNN models and data, thereby motivating a new, important direction in KD methods. 

\noindent \textbf{Model Stealing and KD:} Model stealing involves stealing any information from a DNN model that is desired to be inaccessible to an adversary/end-user. 
Such stealing can happen in multiple ways: \textit{(1) Model Extraction as a Black Box.} An adversary could query existing model-based software, and with just its outputs clones the knowledge into a model of their own; \textit{(2) Using Data Inputs.} An adversary may potentially access similar/same data as the victim, which can be used to extract knowledge/IP; or \textit{Using Model Architecture/Parameters.} An adversary may attempt to extract critical model information -- such as the architecture type or the entire model file  -- through unintentional leaks, academic publications or other means. There have been a few disparate efforts in the past to protect against model/IP stealing in different contexts such as watermark-based methods \cite{watermark1, watermark2}, passport-based methods \cite{passport1, passport2}, dataset inference \cite{di}, and so on. These methods focused on verifying  ownership, while other methods such as \cite{me1, me2} focused on defending against few model extraction attacks. However, the focus of these efforts was different from the one discussed herein. In this work, we specifically explore the recently highlighted problem of KD-based model stealing \cite{nasty, skeptical}. As noted in \cite{nasty,skeptical}, most existing verification and defense methods do not address KD-based stealing, leaving this rather critical problem vulnerable. Our work helps analyze the first defense for KD-based stealing \cite{nasty}, identifies loopholes using simple strategies and also leverages them to propose a newer defense to this problem. We believe our findings will accelerate further efforts in this important space. 
The work closest to ours is one that has been recently published -- Skeptical Student \cite{skeptical} -- which probes the confidentiality of \cite{nasty} by appropriately designing the student (or hacker) architecture. Our approach in this work is different, and focuses on mechanisms of student training, without changing the architecture. 
\footnote{Code available at \url{https://github.com/surgan12/NastyAttacks}.}

\section{Learning from a Nasty Teacher}
\label{sec:method}
\subsection{Background}
\label{sec:background}
\noindent \textbf{\textbf{Knowledge Distillation (KD)}}: KD methods train a smaller student network, $\theta_s$, with the outputs of a typically large pre-trained teacher network, $\theta_t$  alongside the ground-truth labels. Given an input image $\textbf{x}$, the output logits of student given by $\textbf{z}_s= \theta_s(\textbf{x})$ and teacher logits given by $\textbf{z}_t= \theta_t(\textbf{x})$, a temperature parameter $\tau$ is used to soften the logits and obtain a transformed output probability vector using the softmax function: 
\begin{equation}
    \textbf{y}_{s} = \textit{softmax}(\textbf{z}_{s}/\tau), 
    \textbf{y}_{t} = \textit{softmax}(\textbf{z}_{t}/\tau)
\end{equation}
where $\textbf{y}_s$ and $\textbf{y}_t$ are the new output probability vectors of the student and teacher, respectively. The final loss function used to train the student model is given by:
\begin{equation}
    \mathcal{L} = \alpha \cdot \lambda \cdot KL(y_s, y_t) + (1 - \alpha) \cdot \mathcal{L}_{CE}
    \label{eq:hinton-loss}
\end{equation}
where KL stands for Kullback-Leibler divergence, $(\mathcal{L}_{CE})$ represents standard cross-entropy loss, and $\lambda, \alpha$ are two hyperparameters to control the importance of the loss function terms ($\lambda=\tau^{2}$ generally).

\noindent \textbf{KD-based Stealing}: Given a stealer (or student) model, denoted by $\theta_{s}$, and a victim (or teacher) $\theta_{t}$, the stealer is said to succeed in stealing knowledge using KD if by using the input-output information of the victim, it can grasp some additional knowledge which is not accessible in the victim's absence. As stated in \cite{nasty}, this phenomenon can be measured in terms of difference in maximum accuracy of stealer with and without stealing from victim. Formally, stealing is said to happen if:
\begin{equation}
    Acc_{w}(KD(\theta_{s}, \theta_{t})) > Acc_{wo}(\theta_{s})
\end{equation}
where the left expression refers to the accuracy with stealing, and the right one refers to accuracy without stealing.

\noindent \textbf{Defense against KD based Stealing}: Following \cite{nasty}, we consider a method $M$ as defense, if it degrades the student's tendency (or accuracy) of stealing. Formally, considering the accuracy of stealer without defense $M$ as $Acc_{w}(KD(\theta_{s},\theta_{t}))$ and with defense as $Acc_{wm}(KD(M(\theta_{t}, \theta_{t})))$, $ M $ is said to be a defense if:
\begin{equation}
   Acc_{wm}(KD(M(\theta_{s},\theta_{t}))) <  Acc_{w}(KD(\theta_{s},\theta_{t})) 
\end{equation}
\label{sec:nastyteacher}
\noindent \textbf{Nasty Teacher(NT)}\cite{nasty}: The Nasty Teacher methodology transforms the original model to a model which has accuracy as high as the original model (to ensure model usability) but whose output distribution (or logits) significantly camouflages the meaningful information.
\begin{figure}[t]
    \centering
	\includegraphics[width=0.65\linewidth]{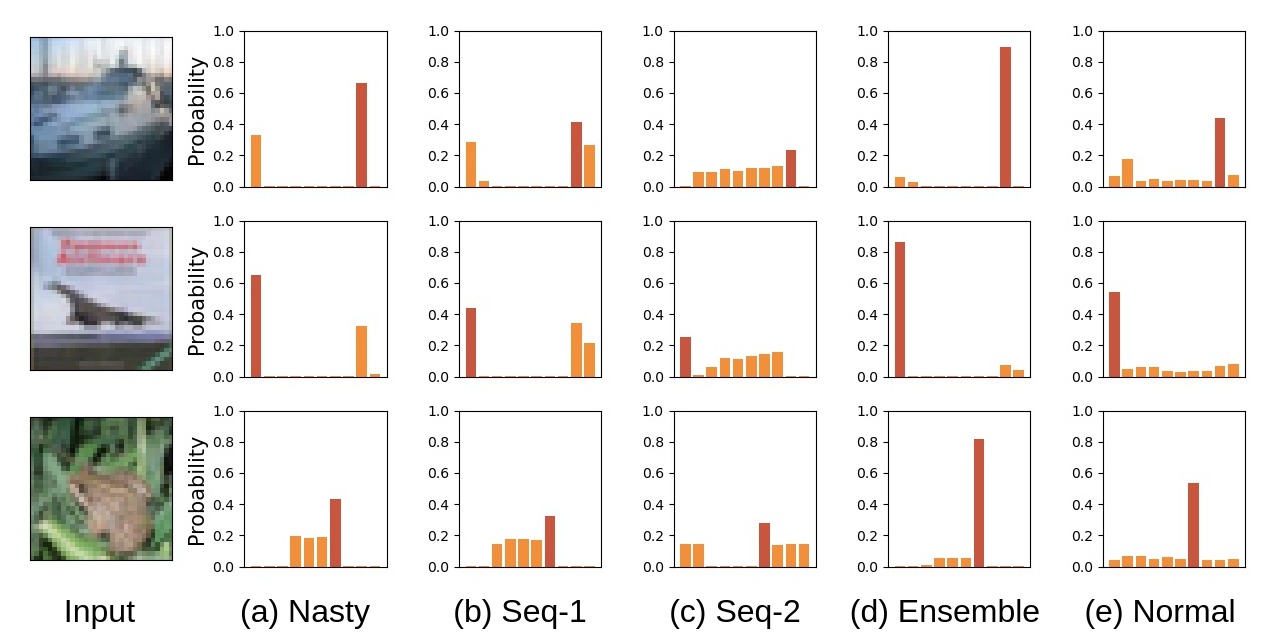}
    \caption{Softmax Outputs of: (a) Nasty Teacher; (b) Seq-1 (Intermediate model $S^{i}$ in Sec \ref{sec:ensmeth}); (c) Seq-2 (Similar to Seq-1); (d) Ensemble of Seq-1 and Seq-2 as used in \ensmeth; and (e) Normal Teacher. Note that the class output distribution of the ensemble is similar to the original (normal) teacher. Maroon = Target class; Orange = other classes.}
    \label{fig:nasty_illustration}
    \vspace{-15pt}
\end{figure}

Formally, given a teacher model $\theta_{t}$, they output a nasty teacher model $\theta_{n}$ trained by minimizing cross-entropy loss $\mathcal{L}_{CE}$ with target labels $y$ (to ensure high accuracy) and also by maximizing KL-Divergence $\mathcal{L}_{KL}$ with the outputs of the original teacher (to maximally contrast or disturb from the original and create a confusing distribution). This can be written as:
\begin{equation}
\label{eq:nasty}
    \mathcal{L}_{n}(\textbf{x},y) = \mathcal{L}_{ce}(\theta_{n}(\textbf{x}), y) - \omega \cdot \tau_{A}^2 \cdot \mathcal{L}_{KL} (\theta_{n}(\textbf{x}), \theta_{t}(\textbf{x}))
\end{equation}
\noindent where $\omega$ is a weighting parameter, and $\tau_{A}^2$ is a temperature parameter. Figure \ref{fig:nasty_illustration}(a) provides a visual illustration of Nasty Teacher's outputs after softmax. We see that when compared to a normal teacher model in Figure \ref{fig:nasty_illustration}(e), it maintains correct class assignments but significantly changes the semantic information contained in the class distribution.
\subsection{Feasibility of KD Based Stealing}
\label{stealing-kd}
As discussed earlier, standard KD techniques \cite{hinton,survey1,survey2,survey3} learn well with just the outputs of the teacher and hence are well-suited for stealing models released as APIs, MLaaS, so on (known as the black-box setting). However, it is also true that the performance of KD methods relies on factors such as training data, architecture choice and the amount of information revealed in the outputs. Thus, one might argue that we can permanently restrict attackers' access to these and prevent KD-based stealing attacks. We now discuss each of these and illustrate the feasibility of KD-based attacks.
\textbf{(1) Restricting Access to Training Data:} While developers try their best to protect such IP assets, as discussed by \cite{skeptical}, there can continue to be numerous reasons for concern: (i) The developers might have bought the data from a vendor who can potentially sell it to others; (ii) Intentionally or not, there is a distinct possibility for data leaks; (iii) Many datasets are either similar to or subsets of large-scale public datasets (ImageNet\cite{imgnet}, BDD100k\cite{bdd100k}, so on), which can be effectively used as proxies; or (iv) Model inversion techniques can be used to recover training data from a pre-trained model \cite{dream2distill, secret_reveal, labonly_inv, bb_frec, GAMIN} both in white-box and black box settings. Such methods \cite{labonly_inv}, in fact, do not even require the soft outputs, just the hard predicted label from the model suffices. Thus, as these methods evolve, we can only expect an adversary to become capable of obtaining training data of sufficient quantity and quality to allow such KD-based stealing.
\textbf{(2) Restricting Access to Architecture:} While developers may not reveal the architecture information completely, common development practices still pose concerns for safety: (i) Most applications simply utilize architectures from existing model hubs \cite{torchhub,tensorflowhub}, sometimes even with the same pre-trained weights (transfer learning), which narrow down the options an adversary needs to try; (ii) Availability of additional tools such as AutoML \cite{automl-example2-learningfeatureengg,automl-example2-oneplayergame}, Neural Architecture Search (NAS) \cite{nas-example1-simpleandefficient,nas-example2-darts} can help attackers search for architectures that match a specific criteria. When combined with the increasingly available compute power (in terms of TPUs, GPUs), this can make models significantly vulnerable to attacks; 
(iii) With advancements in KD methods, it has been shown that knowledge can be distilled from any architecture to the other: \cite{ban} shows distillation with same capacity networks, \cite{tfkd} shows distillation from even poorly trained networks, and so on. Besides, KD has also witnessed techniques that require no data information and still achieve distillation (i.e. data-free distillation \cite{DAFL, dream2distill}). Hence, more opportunities for an attacker to carry out KD-based stealing. 
\textbf{(3) Releasing Incomplete or Randomly Noised Teacher Outputs:} The degree of exposition (i.e. the number of classes revealed) and the clarity of scores (i.e. ease of understanding and inferring from scores) play a vital role in ensuring the quality of knowledge transfer. Hence, one might argue that releasing only top-K class scores can help contain attackers; however, such an approach is infeasible in many use cases where it is necessary to fetch the entire score map. One might also consider adding random noise to make teacher outputs undistillable. This approach generally lowers model performance but also make its failure tracking difficult.\\
\noindent The above discussion motivates the need to develop fundamentally sound strategies to protect against stealing. To this end, we analyze ``Nasty Teacher" \cite{nasty} from two different directions and subsequently leverage insights from KD literature to propose simple yet effective approaches to show that it is still vulnerable to knowledge stealing. We name our attack methods as: \textit{``High Temperature Composition (HTC)"} and \textit{``Sequence of Contrastive Models (SCM)"}, and describe them below. We explain our methods using the Nasty Teacher as the pivot, primarily because it is the only known KD-stealing defense method at this time. Our strategies however are general, and can be applied to any KD based stealing method.
\subsection{High Temperature Composition: \tempmeth}
\label{sec:tempmeth}
\begin{figure}
    \centering

        \includegraphics[width=\linewidth]{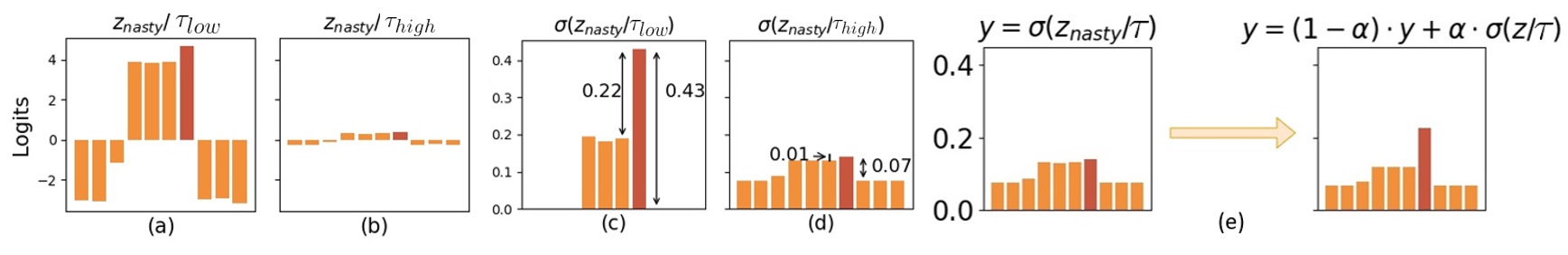}
        \\

    \caption{The effect of temperature in HTC. We demonstrate the logits and probabilities at low temperature (as generally used in KD) and high temperature (as used in HTC). \textbf{(a)} Logits at low temperature $\tau_{low} = 4$, \textbf{(b)} Logits at high temperature $\tau_{high} = 50$, \textbf{(c)} Probabilities at $\tau_{low}$ , \textbf{(d)} Probabilities at $\tau_{high}$, \textbf{(e)} Composing with one-hot to increase the peak.}
    \label{fig:tempmeth-vis}
\end{figure}

\textbf{Motivation: Nasty Teacher Creates Confusing Signals.} In Figure \ref{fig:nasty_illustration}, we observe that the original teacher $\theta_{t}$ emulates a single-peak distribution and consistently has low scores for  incorrect classes. Now, because the Nasty Teacher is trained to contrast with the original teacher, it produces high scores for few incorrect classes, and thus results in a multi-peak distribution (see Fig \ref{fig:nasty_illustration}b). In particular, a few incorrect classes score almost as high as the correct class while other incorrect classes score almost a zero, which, as discussed in \cite{nasty} introduces confusion in the outputs. We attribute this confusion to two key aspects: (i) some low-scoring incorrect classes getting ignored, and (ii) some high-scoring incorrect classes behaving as importantly as the correct class. Fig \ref{fig:tempmeth-vis}c shows a visualization of this observation. 
Since the student model is now forced to learn these incorrect peaks as equally important as the correct class, it gets a false signal and diverges while training.

\noindent {\textbf{Proposition.}} \textit{Transform the output to reduce the degree of confusion in them.}

\noindent {\textbf{Method.}}
We hypothesize that distillation from defenses such as Nasty Teacher can be improved by increasing the relative importance of low-scoring incorrect classes and including their presence in the output. 
Increasing the importance of low-scoring incorrect classes makes the high-scoring incorrect classes lesser important, thus reducing confusion. We note that this idea can be used generically, even independent of the Nasty Teacher's defense. 
To this end, we first soften the teacher's outputs with a high temperature $\tau$ ($\tau$ $>$ 50 in our case). Figure \ref{fig:tempmeth-vis}b shows how this reduces the relative disparity among the scores and brings them closer, thus helping reduce confusion. From the softmax outputs in Figure \ref{fig:tempmeth-vis}d, we further see that this not only allows the other incorrect classes to be viewed in the output but also gives rise to relationships (or variations) which were earlier not visible. Formally, the above operation can be written as: 
\begin{equation}
    \textbf{y}_{nasty} = softmax(\textbf{z}_{nasty} / \tau) 
    \label{eq:y-nasty-htc}
\end{equation}
Although we get a much more informative output, the above transformation does also reduce the relative peak of the correct class, which for distillation may not be ideal. We overcome this by using a convex combination of  $\textbf{y}_{nasty}$ with the one-hot target vector to obtain the final output $\textbf{y}_{net}$, which makes this strategy more meaningful (see Figure \ref{fig:tempmeth-vis})):
\begin{equation}
    \textbf{y}_{net} = (1 - \alpha)  \cdot \textbf{y} + \alpha \cdot \textbf{y}_{nasty}
    \label{eq:y-net-htc}
\end{equation}

\noindent In the above discussion, we propose to create our own training targets which satisfy the two properties to learn despite the Nasty Teacher's defense: (i) has a high peak for the correct class; and (ii) has the rich semantic class score information. Finally, to learn from this teacher and match its distributions, we minimize the cross-entropy lpss between student probabilities ($\textbf{s} = softmax(\textbf{z}_{student})$) and \tempmeth teacher ($\textbf{y}_{net}$) targets as:
\[ \mathcal{L_{HTC}} = - \sum_{i} \textbf{y}_{net, i} \cdot \log{\textbf{s}_i} \]
\noindent Combining the above with Eqn \ref{eq:y-net-htc}, $\mathcal{L}_{CE}$ as cross-entropy loss and $\mathcal{L}_{KL}$ as KL-Divergence, we can now write the above as:
\begin{align}
    \mathcal{L_{HTC}} &= -\sum_{i} ((1 - \alpha) \cdot \textbf{y}_i + \alpha \cdot \textbf{y}_{nasty, i}) \log{\textbf{s}_i} \nonumber \\
    &= -(1 - \alpha)  \cdot \sum_{i} \textbf{y}_i \cdot \log{\textbf{s}_i} - \alpha \cdot \sum_{i} \textbf{y}_{nasty, i} \cdot \log{\textbf{s}_i} \nonumber \\
    &= (1 - \alpha)  \cdot \mathcal{L}_{CE}(\textbf{s}, \textbf{y}) + \alpha \cdot \mathcal{L}_{CE}(\textbf{s}, \textbf{y}_{nasty}) \nonumber \\ \\
    \mathcal{L_{HTC}} &= (1 - \alpha) \cdot \mathcal{L}_{CE}(\textbf{s}, \textbf{y}) + \alpha \cdot \multiplier \cdot \mathcal{L}_{KL}(\textbf{s}, \textbf{y}_{nasty})
    \label{eq:tempmeth-loss}
\end{align}

\begin{figure}[t]
    \centering
	\includegraphics[width=0.8\linewidth]{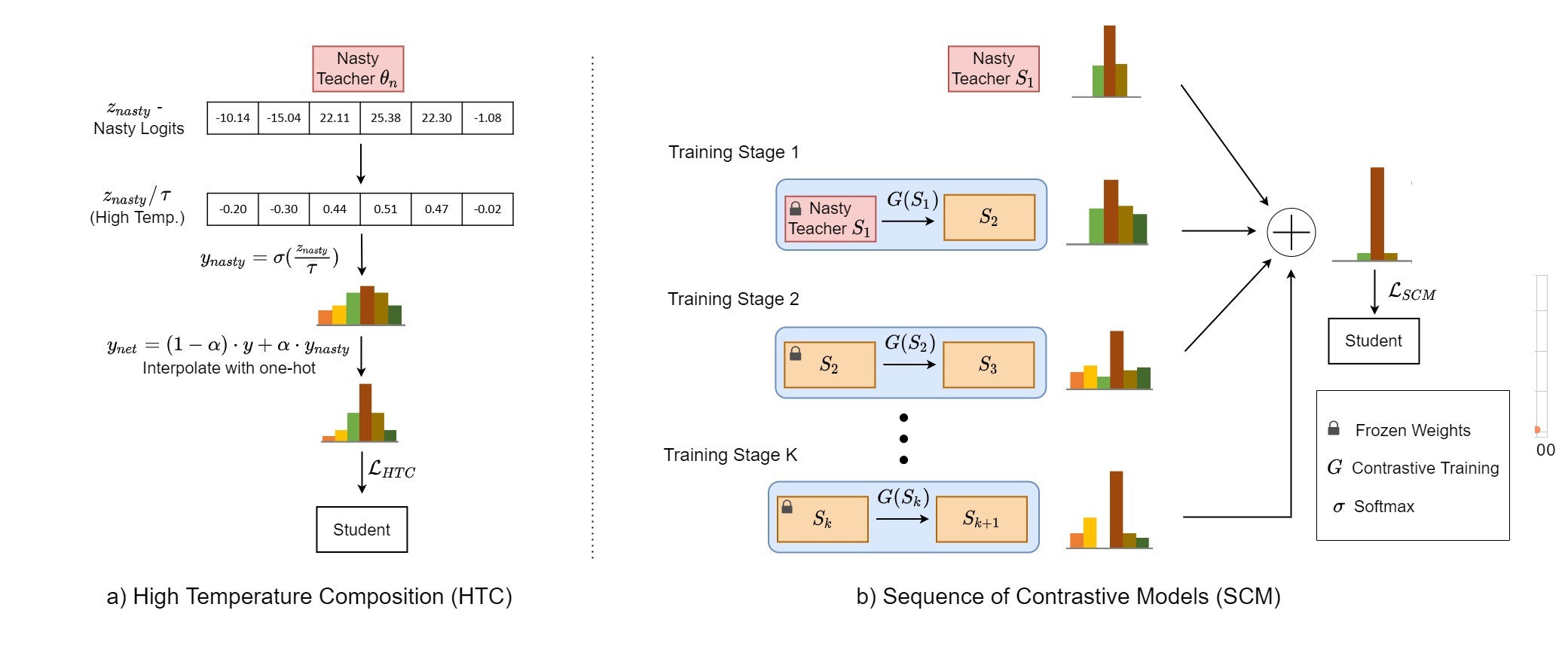}
    \caption{\textit{Illustration of our Methods}: High Temperature Composition (HTC) and Sequence of Contrastive Models (SCM). \textit{HTC} learns from Nasty Teacher (NT) outputs by reducing their confusion, while \textit{SCM} learns by extracting semantic information from them. }
    \label{fig:into-vis}
\end{figure}

\noindent In Equation \ref{eq:tempmeth-loss}, we see that $\mathcal{L}_{CE}$ is replaced with $\mathcal{L}_{KL}$. Here, we use the fact that cross-entropy loss and KL-divergence differ by a term $\sum_{i} \textbf{y}_{nasty,i} \log{\textbf{y}_{nasty,i}}$, which remains a constant for student gradient computation because of fixed teacher outputs $\textbf{y}_{nasty}$. $\mathcal{L}_\mathcal{HTC}$ is thus a KD-loss similar to what was discussed in Sec \ref{sec:background}. To adjust the extent of knowledge transfer, we finally re-weight the KL-divergence term with a hyperparameter multiplier $\multiplier$. Conceptually, $\multiplier$\ does not make a difference to the idea, but we observe this to be useful while training. Figure \ref{fig:into-vis}a provides a visualization for this approach.
\subsection{Sequence of Contrastive Models: \ensmeth}
\label{sec:ensmeth}
\noindent{\textbf{Motivation: Nasty Hides Essential Information.}} As Nasty Teacher selectively causes some incorrect classes to exhibit peaks while inhibiting scores for others, it hides certain inter-class relationships to protect against KD-based attacks.

\noindent {\textbf{Proposition.}}: \textit{Transform the output to extract/recover the essential class relationships or possibly the entire original teacher distribution}

\noindent {\textbf{Method.}}: To begin with, we ask the question - what would happen if we used Eqn \ref{eq:nasty} as is with the given nasty teacher $\theta_{n}$? We would expect to see a model with accuracy as high as the teacher but with presence of class distribution peaks different from it. If we perform this operation for ``k" such sequential stjpg, we can expect to obtain ``k" potentially different output distributions. Building on this thought, we introduce a Sequential Contrastive Training strategy, wherein we form a sequence of ``k" contrastive models by training each model to contrast with the model just before it. Formally, taking the nasty teacher $\theta_{n}$ as the starting point of the sequence, $G$ as the method for generating the next contrastive item in the sequence, which in our approach is the same as the one described in Eqn \ref{eq:nasty}, the sequence $S^{k}$ with $k$ as the sequence length can be written as:
\begin{equation}
S^{k} = \begin{cases} 
       {S}^{i} = \theta_{n} & i=1 \\
       {S}^{i} = G({S}^{i-1}) & 1 < i\leq k
   \end{cases}
\end{equation}
Thus, while maximising KL-divergence between models, each model learns a different probability distribution and has its own unique set of confusing class relationships. We then take an ensemble of their outputs, which is denoted as $\textbf{z}_{ens}$ and use this alongside the ground truth labels $y$ to train the student (or stealer) model, i.e.
\begin{multline}
    \mathcal{L_{SCM}} = (1 - \alpha) \cdot \mathcal{L}_{CE} (softmax(\textbf{z}_{s}), \textbf{y}) \\
    + \alpha \cdot \multiplier \cdot \mathcal{L}_{KL} (softmax(\textbf{z}_{s}/\tau), softmax(\textbf{z}_{ens}/\tau))
\end{multline} 
where hyperparameters $\multiplier$, $\alpha$, $\tau$ have the same meaning as in Eqn \ref{eq:tempmeth-loss}.
The core intuition behind \ensmeth is that a relationship may be important in its own distribution but only the essential relationships will be important across distributions. Therefore, by taking the ensemble, we not only capture such relationships but also obtain a distribution closer to the original teacher. This idea is visualized in Figure \ref{fig:nasty_illustration} where (b) and (c) represent two successive checkpoints in the sequence, and (d) represents their ensemble. It can be clearly noted that different items in the sequence consistently output diversity in class relationships, and further, their ensemble illustrates a class distribution closer to the original teacher (Figure \ref{fig:nasty_illustration}e).
\section{Experiments and Results}
\label{sec:results}
\subsection{Experimental Setup}
\label{sec:setup}
\noindent For all the experiments, we follow the same models and training configurations as used in the code provided by \cite{nasty}\footnote{https://github.com/VITA-Group/Nasty-Teacher}.

\noindent \textbf{Datasets and Network Baselines:} For datasets, we use CIFAR-10 (C10), CIFAR-100 (C100) and TinyImageNet (TIN) datasets in our evaluation. For teacher (or victim) networks, we use ResNet18 \cite{resnet} for CIFAR-10, both Resnet18 and ResNet50 for CIFAR-100 and TinyImageNet. For student (or stealer) networks, we use MobileNetv2 \cite{mobilev2}, ShuffleNetV2 \cite{shufflev2}, and Plain CNNs \cite{takd} in CIFAR-10 and CIFAR-100 and MobileNetV2, ShuffleNetV2 in TinyImageNet. For baselines, Vanilla in Table \ref{tab:main} refers to the cross entropy training, \textit{Normal KD}/\textit{Nasty KD} refer to the distillation (or stealing) with original/nasty teacher, \textit{Skeptical} refers to the recent work \cite{skeptical}, and \textit{HTC, SCM} refers to our approaches. 

\noindent \textbf{Training Baselines:} We follow the parameter choice from \cite{nasty}. For generating Nasty Teacher in Eq. \ref{eq:nasty}, $\tau_{A}$ is set to 4 for CIFAR-10 and 20 for CIFAR-100/TinyImageNet, correspondingly $\omega$ to 0.04, 0.005, 0.01 for CIFAR-10, CIFAR-100, TinyImageNet respectively. For both Nasty KD and Normal KD, ($\alpha$, $\tau$) parameters of distillation in Eq. \ref{eq:hinton-loss} are set to (0.9, 4) in plain CNNs and (0.9, 20) in MobileNetV2, ShuffleNetV2. Plain CNNs are optimized with Adam\cite{adam} (LR=1e-3) for 100 epochs while MobileNetV2, ShuffleNetV2 are optimized with SGD (momentum=0.9, weight decay=5e-4, initial LR=0.1) for 160 epochs with LR decay of 10 at epoch [80, 120] in CIFAR-10 and for 200 epochs with LR decay of 5 at epoch [60, 120, 160] in CIFAR-100. Training settings in TinyImageNet are same as used in CIFAR-100. Moreover, all the experiments use batch size of 128 with standard image augmentations.

\noindent \textbf{Training \tempmeth and \ensmeth:} For both \tempmeth and SCM, we search $\multiplier$\ in \{1, 5, 10, 50\}, $\tau$ in \{4, 20, 50, 100\}, $\alpha$ in \{0.1,....0.9\}. For TinyImageNet, we include $\tau=200$ also in the earlier set of $\tau$. While any number of sequence models can conceptually be chosen for \ensmeth, our experiments only leverage a max. of 4 such Sequence Contrastive models. Note, for all these we choose the search space rather intuitively and finally present their impact in Sec. \ref{sec:ablations}.\\
\subsection{Quantitative Results}
\noindent We report the results in Table \ref{tab:main} and clearly observe the degradation of student performance while learning from Nasty teacher. Subsequently, our approaches: \tempmeth and \ensmeth increase learning from Nasty by upto 58.75\% in CIFAR-10, 68.63\% in CIFAR-100 and 60.16\% in TinyImageNet. We significantly outperform the recent state of the art Skeptical \cite{skeptical} and unlike them we also consistently outperform the Vanilla training. Moreover, many times we achieve very close performances and other few times even better performance than Normal-Teacher (i.e stealing from unprotected victim model), see the $\ssymbol{4}$ marked cells in table \ref{tab:main}. Further, we combine our training methods \tempmeth and \ensmeth with the novel stealing architecture Skeptical Student \cite{skeptical} and report improvements in Table \ref{tab:main}.
\begin{table}[t]
    
    \centering
    
    \begin{tabular}{ccc cc|cccc H | cc}
    \hline
    \hline
        Dataset & Tch. & Stu. & Vanilla & Normal & Nasty & Skep. & \tempmeth & \ensmeth & $\Delta$ & Skep. & Skep. \\
         & & & & KD & KD & (NeurIPS'21) & & & & +\tempmeth & +\ensmeth \\
        \hline
        \hline
        \multirow{3}{*}{C10} & \multirow{3}{*}{Res18} & CNN & 86.31 & 87.83 & 82.27 & 86.71  & \underline{87.38} & \textbf{87.85} & +1.14 & 87.17 & 87.04 \\
         & & Mob & 89.58 & 89.30 & 31.73 & 90.53  & 90.03$^\ssymbol{4}$ & \underline{90.48}$^\ssymbol{4}$ & -0.05 & \underline{91.45}$^\ssymbol{4}$ & \textbf{91.55}$^\ssymbol{4}$ \\
         & & Shuf & 91.03 & 91.17 & 79.73 & 91.34 & 91.61$^\ssymbol{4}$ & 91.93$^\ssymbol{4}$ & +0.59 & \underline{92.45}$^\ssymbol{4}$ & \textbf{92.76}$^\ssymbol{4}$ \\
         \hline
        \multirow{3}{*}{C100} & \multirow{3}{*}{Res18} & CNN & 58.38 & 62.35 & 58.62  & 58.38 & \underline{61.21} & \textbf{61.31} & +2.39 & 59.64 & 59.17 \\
         & & Mob & 68.90 & 72.75 & 3.15 & 66.89 & 71.01 & 71.06 & +4.17 & \underline{71.48} & \textbf{71.74} \\
         & & Shuf & 71.43 & 74.43 & 63.67 & 70.00 & 74.04 & \textbf{75.23}$^\ssymbol{4}$ & +5.23 & 73.78 & \underline{74.23} \\
         \hline
        \multirow{3}{*}{C100} & \multirow{3}{*}{Res50} & CNN & 58.38 & 61.84 & 58.93 & 59.15 & \textbf{61.24} & \underline{59.58}  & +2.09 & 59.48 & 59.17 \\
         & & Mob & 68.90 & 72.22 & 3.03 & 66.65 & 70.49 & \textbf{71.66} & +5.01 & \underline{71.54} & 71.16 \\
         & & Shuf & 71.43 & 73.91 & 62.8 & 70.02 & 72.37 & 73.25 & +3.23 & \underline{73.60} & \textbf{74.73}$^\ssymbol{4}$ \\
        \hline
        \multirow{2}{*}{TIN} & \multirow{2}{*}{Res18} & Mob & 55.69 & 61.00 & 0.85 & 47.37 & 56.28 & \textbf{61.01}$^\ssymbol{4}$ & +13.64 & \underline{59.05} & 58.88 \\
         & & Shuf & 60.30 & 63.45 & 23.78 & 54.78 & 60.46 & 62.09 & +7.31 & \textbf{63.05} & \underline{62.28} \\
        \hline
        \multirow{2}{*}{TIN} & \multirow{2}{*}{Res50} & Mob & 55.89 & 57.84 & 1.10 & 48.21 & 56.00 & \textbf{59.46}$^\ssymbol{4}$ & +11.25 & 58.56$^\ssymbol{4}$ & \underline{58.88}$^\ssymbol{4}$ \\
         & & Shuf & 60.30 & 62.02 & 24.27 & 56.08 & 60.80 & 61.55 & +5.47 & \textbf{63.14}$^\ssymbol{4}$ & \underline{61.92} \\
        \hline
        \hline
    \end{tabular}
    \caption{{\textit{Accuracy (higher is better)}} of \tempmeth and \ensmeth against baselines. \textbf{bold} represents best performance, \underline{underline} the second best in learning from Nasty Teacher.  {$\ssymbol{4}$ \textit{represents instances}} that even outperform the Normal KD. {\textit{Abbreviations}} -- Tch : Teacher, Stu : Student, Skep : Skeptical \cite{skeptical}, Res50 : ResNet50, Res18 : ResNet18, Mob : MobileNetV2, Shuf : ShuffleNetV2.}
    \label{tab:main}
\end{table}


\begin{table}[t]

    \begin{minipage}{0.5\linewidth}
    \centering 
    \begin{tabular}{ccccc}
    \hline
    \hline
         & Vanilla  & Nasty KD  & Ours \\
    \hline
        CNN & 58.38  & 58.62 & \textbf{61.25}  \\
        ShuffleNetV2 & 71.43 & 63.67  & \textbf{72.71} \\
    \hline
    \hline
    \end{tabular}
    \caption*{(a) Effect of architecture choice.}
    \end{minipage}
    \begin{minipage}{0.5\linewidth}
    \centering
    \begin{tabular}{ccccc}
    \hline
    \hline
         & 2 & 3 & 4 \\
    \hline
        CNN & 60.99 & 61.24 & \textbf{61.31} \\
        ShuffleNetV2 & 74.45 & 74.24 & \textbf{75.23} \\
    \hline
    \hline
    \end{tabular}
    \caption*{(b) Effect of sequence length.}
    
    \end{minipage}

\caption{Analysing hyperparameter choice for SCM.}
\label{tab:hyp-ens}

\end{table}
\subsection{Ablation Studies}

\label{sec:ablations}

\noindent{\textbf{Choice of $\tau$, $\multiplier$ in HTC:}} \tempmeth (Sec 3.3) depends on softening temperature ($\tau$) which adjusts the optimal representation, and multiplier $\multiplier$\ which adjusts the optimal weight to transfer this knowledge. Thus, we now study each of these parameters separately. We also note that $\alpha$ primarily controls the ground truth signal, hence, we omit varying $\alpha$ here and set it to 0.9. We vary the $\tau$ as 10, 20, 50, 100 with $\multiplier$\ = 50 on ShuffleNetV2 and present results in Fig. \ref{fig:ablations-temp} (a),(b) to demonstrate the effect of setting temperature to an optimal higher value. We then vary $\multiplier$\ keeping the $\tau$ at 50 to observe the motivation of its careful selection in \ref{fig:ablations-temp}(d).  \begin{wraptable}{r}{0.6\linewidth}
    \centering
    \begin{tabular}{cc|cc}
    \hline
     & CIFAR10 & CIFAR100 & CIFAR100 \\
     & ResNet18 & ResNet18 & ResNet50 \\
     \hline
    Nasty Teacher & 0.4446 & 14.6363 & 10.4892 \\
    SCM Ensemble & \textbf{0.2471} &\textbf{ 5.8398 }&\textbf{ 5.3165} \\
    \hline
    \end{tabular}
    \caption{KL-Divergence scores of a given model (col 1) with the original teacher. }
    \label{tab:ens-kl-div}
\end{wraptable} 

\noindent \noindent\textbf{Choice of Architecture and $k$ in \ensmeth:} Given a Nasty Teacher, we currently use the same architecture type to generate ${S}^{1}, {S}^{2},..{S}^{k}$. However, we now explore if \ensmeth generalizes to different architecture choices in Table \ref{tab:hyp-ens}(a). Rather than obtaining seq. items as ResNet18 $\rightarrow$ ResNet18 i.e ${S^{i}}_{ResNet18}=G({S^{i-1}}_{ResNet18})$, we hereby do ${S^{i}}_{ShuffleNetV2}=G({S^{i-1}}_{ResNet18})$, and finally ensemble generated ShuffleNetV2 ($S^{2})$ and original ResNet18 ($S^{1}$) to train the students (or stealers). Table \ref{tab:hyp-ens} further shows that even with different architecture, \ensmeth can extract information from Nasty Teacher. Moving ahead, we now ablate on \textit{sequence length $k$}. From Table \ref{tab:hyp-ens}(b), we observe improvement in performance as we increase number of sequence items. This can be intuitively linked to getting better estimate of essential relationships with more number of models, hence improved distillation. \\

\noindent \noindent\textbf{Similarity of \ensmeth and Original Teacher:} As with SCM we seek to potentially recover the original teacher distribution, thus in addition to the visualization we present in \ref{fig:nasty_illustration} we hereby conduct this study to quantitatively estimates the closeness of the SCM ensemble and normal teacher. We use KL-Divergence as our metric and present results in Table \ref{tab:ens-kl-div}. It can can be clearly seen that compared to Nasty Teacher, SCM ensemble consistently results in low KL-Divergence, thereby lending support to our motivation in recovering original teacher. \\

\subsection{Limited and No Data Setting} 
In this section, we consider settings to learn from the nasty teacher by using only a part of the data. Specifically, we experiment three settings and for the sake of simplicity use \tempmeth to investigate learning from nasty under these settings.

\noindent \textbf{No Label Available:} Here, we consider no access to the labels for data used in \ref{sec:setup}. Table \ref{tab:nolabel-results} includes the results against this setting. 

\begin{figure}[t]
\centering
    \includegraphics[width=0.8\linewidth]{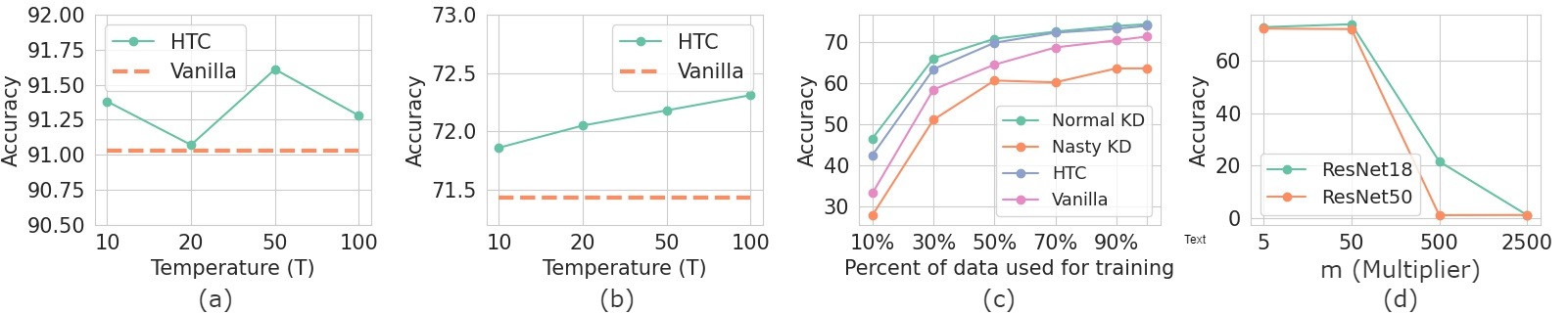}
    \caption{Effect of Temperature: \textbf{(a)} C10, Student : ShuffleNetV2, Teacher : ResNet18 and \textbf{(b)} C100, Student : ShuffleNetV2, Teacher : ResNet50. \textbf{(c)} Effect of percentage of data. \textbf{(d)} Effect of multiplier $m$.}
    \label{fig:ablations-temp}
\end{figure}

\noindent \textbf{Limited Data Available:} Here, we evaluate the learning by varying the percent of data used [10\%, 30\%, 50\%, 70\%, 90\%]. From Fig. \ref{fig:ablations-temp}(c), we observe to better extract knowledge in all data-subset fractions. Specifically, performance difference becomes even more notable when very low fraction is used (like in 10\% setting, we perform approx. 15\% better than Nasty Teacher). In addition, we also observe that we always remains significantly closer to the original teacher. 

\begin{table}
    \begin{minipage}[c]{0.45\linewidth}
        \centering
        \begin{tabular}{ccc}
            \hline
            \hline
            Network & Nasty & HTC \\
            \hline
            CNN & 16.38 & \textbf{26.03} \\
            MobileNetV2 & 41.79 & \textbf{52.96} \\
            ShuffleNetV2 & 70.04 & \textbf{76.29} \\
            \hline
        \end{tabular}
        \caption{\label{tab:datafree-results} No Data Available Results for CIFAR10.}
    \end{minipage}
    \begin{minipage}[c]{0.55\linewidth}
        \centering
        \begin{tabular}{cccc}
            \hline
            \hline
            Dataset & Network & Nasty & HTC \\
            \hline
            CIFAR-10 & CNN & 82.64 & \textbf{87.48} \\
            CIFAR-100 & Shufflenetv2 & 64.41 & \textbf{74.17} \\
            \hline
        \end{tabular}
        \caption{\label{tab:nolabel-results} No Label Available results. Using images without their ground truth.}        
    \end{minipage}
\end{table}
\noindent \textbf{No Data Available:} Here, we consider the setting where no data is available. More precisely, we take the existing data-free distillation method \cite{DAFL}, and add our method to it. Table \ref{tab:datafree-results} demonstrates consistent improvement while learning from nasty in this setting.

\subsection{Leveraging SCM for an improved defense}
\noindent Previously, we discuss the vulnerabilities of Nasty Teacher \cite{nasty}. We now discuss this section in regards to improve ``Nasty Teacher" defense.



\noindent Nasty Teacher derives its efficacy from the peaks created for the incorrect classes 
\begin{wrapfigure}{r}{0.5\textwidth}
    \begin{tabular}{cc}
        \includegraphics[width=0.4\linewidth]{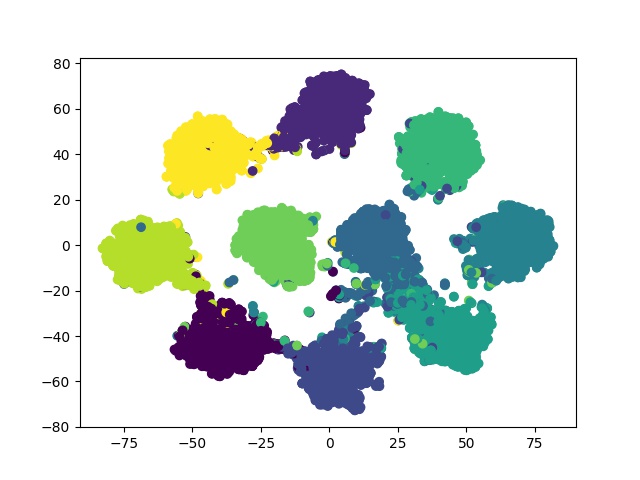} & 
        \includegraphics[width=0.4\linewidth]{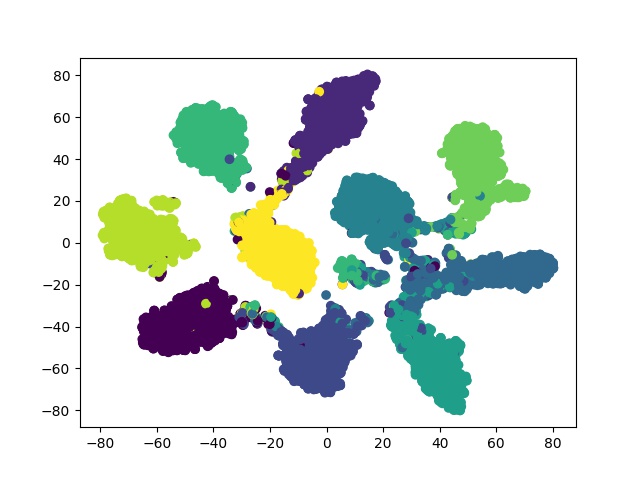} \\
        (a) & (b) \\
        \includegraphics[width=0.4\linewidth]{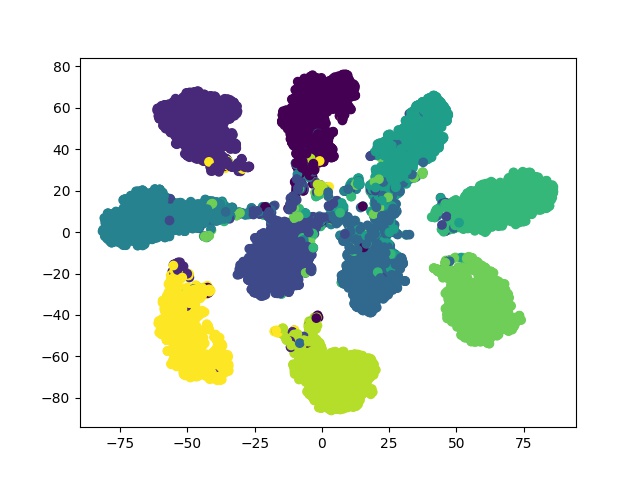}
        &  \includegraphics[width=0.4\linewidth]{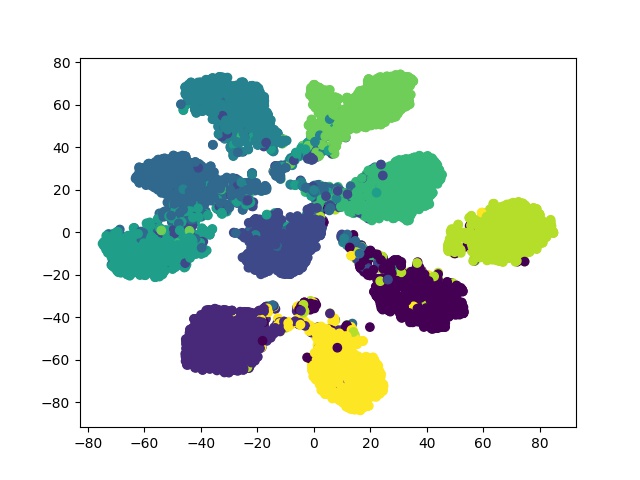} \\
        (c) & (d) \\
    \end{tabular}
    \caption{t-SNE plots of features before FC layers for \ensmeth on CIFAR-10. Model : ResNet-18, \textbf{a)} Normal Teacher \textbf{b)} Nasty Teacher \textbf{c)} Seq-1 \textbf{d)} Seq-2}
    \label{fig:tsne-scm}
\end{wrapfigure}
(Fig. \ref{fig:nasty_illustration}). 
Along the lines that each in-correct peak adds to the confusion while learning, we hypothesize that the number of in-correct peaks in the output distribution affects the effectiveness of the defense. In general, more number of peaks can be correlated with the increased confusion, hence a better defense. However, in case of Nasty we often see the number of peaks to be not many in comparison to the total number of classes (see Fig. \ref{fig:nasty_illustration}(a) for illustration), thereby creating a chance to improve Nasty's defense via increasing the number of in-correct peaks. In lieu of this observation, we propose to use our previously proposed Sequence of Contrastive Model Training for an improved defense. As model obtained with each SCM's step displays diverse set of peaks, we expect them to exhibit more number of peaks, largely because they were trained in a way to eventually contrast from nasty teacher (i.e low number of peaks). Fig. \ref{fig:nasty_illustration} illustrates this via columns (b),(c) where Seq-1/Seq-2 to have more number of peaks than Nasty Teacher. Thus, we choose one of these intermediate sequence models ($S^{1}, S^{2}...S^{k}$) for teacher and further show our results (denoted by \textbf{Ours}) in Table \ref{tab:defense}. For ease of our exploration, we typically test out with first or second model from our sequence in \ref{sec:ensmeth} as our defense teacher. Though our model incurs a small drop in accuracy ($<2\%$), the significant improvement (sometimes, as much as $~10\%$ ) it produces while defending makes it attractive for application.  
Diving deep into the results, one may consider this degradation in KD related to the decrease in accuracy. We now evaluate this dimension for a better understanding. Specifically, we obtain a model with an accuracy similar to ours by training a Nasty Teacher (refer \ref{sec:background}) followed by early stopping it at the desired accuracy. We dub this approach Nasty KD ES (Early Stopped) and include the results in Table \ref{tab:defense}. We observe that for a similar accuracy Nasty KD ES has a significantly poor defense against \textbf{our method}, and in some cases it even defends poorly compared to original Nasty KD. Moreover, we visualize the features of intermediate sequence models in Fig. \ref{fig:tsne-scm} t-SNE \cite{tsne} plots and infer the intermediates i.e Seq-1 and Seq-2 to possess a similar class separation as the normal teacher \ref{fig:tsne-scm}(a) while maintaining the desired defense.
\begin{table}[t]
    \centering
    \begin{tabular}{cccc|c}
    \hline
    \hline
        CIFAR-10 & Normal Teacher &  Nasty KD  & Nasty KD ES  & Ours \\
        ResNet18 & (Acc=95.09) & (Acc=94.28) & (Acc=93.02) & (Acc=92.99) \\
        \hline
        CNN & 85.99 & \underline{82.27}  & 84.62 & \textbf{80.13}\\
        MobileNetV2 & 89.58  & 31.73 &  \underline{28.58} & \textbf{22.12}  \\
        ShuffleNetV2 & 91.03 & \underline{79.73}  & 87.81  & \textbf{73.23}\\
    \hline
    \hline
    \hline
        CIFAR-100 & Normal Teacher &  Nasty KD  & Nasty KD ES & Ours \\
        ResNet50 & (Acc=77.96)  & (Acc=77.4) & (Acc=76.2) & (Acc=75.44) \\
    \hline
        CNN & 58.38 & 58.93 & \underline{58.45} & \textbf{54.26} \\
        MobileNetV2 & 71.43 & 3.03 & \underline{2.16} & \textbf{1.4} \\
        ShuffleNetV2 & 68.90  & 63.16 & \underline{62.95} & \textbf{ 54.00}\\
        \hline
        \hline
    \end{tabular}
 \caption{Results of our defense, Nasty KD and Nasty KD ES. Training hyperparameters are same as discussed in section 
 \ref{sec:setup}.}
    \label{tab:defense}
\end{table}

\section{Conclusion}
In this work, we focus on the threat of KD-based model stealing; specifically, the recent work Nasty Teacher \cite{nasty} which proposes a defense against such stealing. We study \cite{nasty} from two different directions and systematically show that we can still extract knowledge from Nasty Teacher with our approaches: HTC and SCM. Extensive experiments demonstrate our efficacy to extract knowledge from Nasty. Leveraging the insights we gain in our approaches, we finally also discuss an extension of Nasty Teacher that serves as a better defense. Concretely, we highlight a few dimensions that affect the defense against KD-based stealing to facilitate subsequent efforts in this direction. As our future work, we intend to improvise the existing defense and simultaneously explore such stealing in a relatively white-box setting, wherein the goal will be to defend even if the adversary gets hold of the model features/parameters.
\\
\textbf{Acknowledgements.} This work was partly supported by the Department of Science and Technology, India through the DST ICPS Data Science Cluster program. We also thank the anonymous reviewers for their valuable feedback in improving the presentation of this paper.

\clearpage
\bibliographystyle{splncs04}
\bibliography{./eccv2022submission.bbl}

\begin{thebibliography}{10}
\providecommand{\url}[1]{\texttt{#1}}
\providecommand{\urlprefix}{URL }
\providecommand{\doi}[1]{https://doi.org/#1}

\bibitem{torchhub}
Pytorch model hub (last accessed on 8 march 2022).
  \url{https://pytorch.org/hub/} (2022)

\bibitem{tensorflowhub}
Tensorflow model hub (last accessed on 8 march 2022).
  \url{https://www.tensorflow.org/hub} (2022)

\bibitem{GAMIN}
A{\"i}vodji, U., Gambs, S., Ther, T.: Gamin: An adversarial approach to
  black-box model inversion. ArXiv  (2019)

\bibitem{bucilua}
Bucilua, C., Caruana, R., Niculescu-Mizil, A.: Model compression. In:
  Proceedings of the 12th ACM SIGKDD International Conference on Knowledge
  Discovery and Data Mining. p. 535–541. KDD '06, Association for Computing
  Machinery (2006)

\bibitem{DAFL}
Chen, H., Wang, Y., Xu, C., Yang, Z., Liu, C., Shi, B., Xu, C., Xu, C., Tian,
  Q.: Dafl: Data-free learning of student networks. In: ICCV (2019)

\bibitem{delange2021pami-cl-survey}
Delange, M., Aljundi, R., Masana, M., Parisot, S., Jia, X., Leonardis, A.,
  Slabaugh, G., Tuytelaars, T.: A continual learning survey: Defying forgetting
  in classification tasks. IEEE Transactions on Pattern Analysis and Machine
  Intelligence  (2021)

\bibitem{nas-example1-simpleandefficient}
Elsken, T., Metzen, J.H., Hutter, F.: Simple and efficient architecture search
  for convolutional neural networks (2018)

\bibitem{passport1}
Fan, L., Ng, K.W., Chan, C.S.: Rethinking deep neural network ownership
  verification: Embedding passports to defeat ambiguity attacks. In: Advances
  in Neural Information Processing Systems. vol.~32. Curran Associates, Inc.
  (2019)

\bibitem{ban}
Furlanello, T., Lipton, Z.C., Tschannen, M., Itti, L., Anandkumar, A.:
  Born-again neural networks. In: International Conference on Machine Learning,
  {ICML} 2018. vol.~80, pp. 1602--1611 (2018)

\bibitem{automl-example2-oneplayergame}
Gaudel, R., Sebag, M.: Feature selection as a one-player game. p. 359–366.
  ICML'10, Omnipress, Madison, WI, USA (2010)

\bibitem{ard}
Goldblum, M., Fowl, L., Feizi, S., Goldstein, T.: Adversarially robust
  distillation. Proceedings of the AAAI Conference on Artificial Intelligence
  \textbf{34}(04),  3996--4003 (Apr 2020)

\bibitem{survey1}
Gou, J., Yu, B., Maybank, S.J., Tao, D.: Knowledge distillation: A survey.
  International Journal of Computer Vision  \textbf{129}(6),  1789--1819 (Jun
  2021)

\bibitem{resnet}
{He}, K., {Zhang}, X., {Ren}, S., {Sun}, J.: Deep residual learning for image
  recognition. In: 2016 IEEE Conference on Computer Vision and Pattern
  Recognition (CVPR). pp. 770--778 (2016)

\bibitem{hinton}
Hinton, G., Vinyals, O., Dean, J.: Distilling the knowledge in a neural
  network. In: NIPS Deep Learning and Representation Learning Workshop (2015)

\bibitem{me2}
Juuti, M., Szyller, S., Dmitrenko, A., Marchal, S., Asokan, N.: Prada:
  Protecting against dnn model stealing attacks. 2019 IEEE European Symposium
  on Security and Privacy (EuroS\&P) pp. 512--527 (2019)

\bibitem{labonly_inv}
Kahla, M., Chen, S., Just, H.A., Jia, R.: Label-only model inversion attacks
  via boundary repulsion (To Appear CVPR 2022)

\bibitem{me1}
Kariyappa, S., Qureshi, M.K.: Defending against model stealing attacks with
  adaptive misinformation. In: Proceedings of the IEEE/CVF Conference on
  Computer Vision and Pattern Recognition (CVPR) (June 2020)

\bibitem{adam}
Kingma, D.P., Ba, J.: Adam: {A} method for stochastic optimization. In: Bengio,
  Y., LeCun, Y. (eds.) 3rd International Conference on Learning
  Representations, {ICLR} 2015, San Diego, CA, USA, May 7-9, 2015, Conference
  Track Proceedings (2015)

\bibitem{skeptical}
{Kundu}, S., {Sun}, Q., {FU}, Y., {Pedram}, M., {Beerel}, P.A.: Analyzing the
  confidentiality of undistillable teachers in knowledge distillation. In: 35th
  Neural Information Processing Systems (2021)

\bibitem{nas-example2-darts}
Liu, H., Simonyan, K., Yang, Y.: {DARTS}: Differentiable architecture search.
  In: International Conference on Learning Representations (2019)

\bibitem{survey3}
Liu, Y., Zhang, W., Wang, J., Wang, J.: Data-free knowledge transfer: A survey
  (2021)

\bibitem{nasty}
Ma, H., Chen, T., Hu, T.K., You, C., Xie, X., Wang, Z.: Undistillable: Making a
  nasty teacher that {\{}cannot{\}} teach students. In: International
  Conference on Learning Representations (2021)

\bibitem{shufflev2}
Ma, N., Zhang, X., Zheng, H.T., Sun, J.: Shufflenet v2: Practical guidelines
  for efficient cnn architecture design. In: Proceedings of the European
  Conference on Computer Vision (ECCV) (September 2018)

\bibitem{tsne}
van~der Maaten, L., Hinton, G.: Visualizing data using {t-SNE}. Journal of
  Machine Learning Research  \textbf{9},  2579--2605 (2008)

\bibitem{di}
Maini, P., Yaghini, M., Papernot, N.: Dataset inference: Ownership resolution
  in machine learning. In: International Conference on Learning Representations
  (2021)

\bibitem{takd}
Mirzadeh, S., Farajtabar, M., Li, A., Levine, N., Matsukawa, A., Ghasemzadeh,
  H.: Improved knowledge distillation via teacher assistant. Proceedings of the
  AAAI Conference on Artificial Intelligence  \textbf{34},  5191--5198 (04
  2020)

\bibitem{automl-example2-learningfeatureengg}
Nargesian, F., Samulowitz, H., Khurana, U., Khalil, E.B., Turaga, D.: Learning
  feature engineering for classification. In: Proceedings of the 26th
  International Joint Conference on Artificial Intelligence. p. 2529–2535.
  IJCAI'17, AAAI Press (2017)

\bibitem{rkd}
Park, W., Kim, D., Lu, Y., Cho, M.: Relational knowledge distillation. In:
  Proceedings of the IEEE Conference on Computer Vision and Pattern
  Recognition. pp. 3967--3976 (2019)

\bibitem{fskd}
Rajasegaran, J., Khan, S., Hayat, M., Khan, F.S., Shah, M.: Self-supervised
  knowledge distillation for few-shot learning.
  https://arxiv.org/abs/2006.09785  (2020)

\bibitem{bb_frec}
Razzhigaev, A., Kireev, K., Kaziakhmedov, E., Tursynbek, N., Petiushko, A.:
  Black-box face recovery from identity features. In: ECCV Workshops (2020)

\bibitem{fitnet}
Romero, A., Ballas, N., Kahou, S.E., Chassang, A., Gatta, C., Bengio, Y.:
  Fitnets: Hints for thin deep nets. In: Bengio, Y., LeCun, Y. (eds.)
  International Conference on Learning Representations, {ICLR} 2015

\bibitem{imgnet}
Russakovsky, O., Deng, J., Su, H., Krause, J., Satheesh, S., Ma, S., Huang, Z.,
  Karpathy, A., Khosla, A., Bernstein, M., Berg, A.C., Fei-Fei, L.: {ImageNet
  Large Scale Visual Recognition Challenge}. International Journal of Computer
  Vision (IJCV)  \textbf{115}(3),  211--252 (2015)

\bibitem{mobilev2}
Sandler, M., Howard, A.G., Zhu, M., Zhmoginov, A., Chen, L.C.: Mobilenetv2:
  Inverted residuals and linear bottlenecks. 2018 IEEE/CVF Conference on
  Computer Vision and Pattern Recognition pp. 4510--4520 (2018)

\bibitem{watermark1}
Uchida, Y., Nagai, Y., Sakazawa, S., Satoh, S.: Embedding watermarks into deep
  neural networks. In: Proceedings of the 2017 ACM on International Conference
  on Multimedia Retrieval. p. 269–277 (2017)

\bibitem{survey2}
Wang, L., Yoon, K.: Knowledge distillation and student-teacher learning for
  visual intelligence: A review and new outlooks. IEEE Transactions on Pattern
  Analysis \& Machine Intelligence (01), ~1--1 (5555)

\bibitem{fsp}
{Yim}, J., {Joo}, D., {Bae}, J., {Kim}, J.: A gift from knowledge distillation:
  Fast optimization, network minimization and transfer learning. In: 2017 IEEE
  Conference on Computer Vision and Pattern Recognition (CVPR). pp. 7130--7138
  (2017)

\bibitem{dream2distill}
Yin, H., Molchanov, P., Alvarez, J.M., Li, Z., Mallya, A., Hoiem, D., Jha,
  N.K., Kautz, J.: Dreaming to distill: Data-free knowledge transfer via
  deepinversion. In: Proceedings of the IEEE/CVF Conference on Computer Vision
  and Pattern Recognition (CVPR) (June 2020)

\bibitem{bdd100k}
Yu, F., Chen, H., Wang, X., Xian, W., Chen, Y., Liu, F., Madhavan, V., Darrell,
  T.: Bdd100k: A diverse driving dataset for heterogeneous multitask learning.
  In: IEEE/CVF Conference on Computer Vision and Pattern Recognition (CVPR)
  (June 2020)

\bibitem{tfkd}
Yuan, L., Tay, F.E., Li, G., Wang, T., Feng, J.: Revisiting knowledge
  distillation via label smoothing regularization. In: Proceedings of the
  IEEE/CVF Conference on Computer Vision and Pattern Recognition. pp.
  3903--3911 (2020)

\bibitem{at}
Zagoruyko, S., Komodakis, N.: Paying more attention to attention: Improving the
  performance of convolutional neural networks via attention transfer. In:
  International Conference on Learning Representations, {ICLR} 2017

\bibitem{watermark2}
Zhang, J., Chen, D., Liao, J., Fang, H., Zhang, W., Zhou, W., Cui, H., Yu, N.:
  Model watermarking for image processing networks. Proceedings of the AAAI
  Conference on Artificial Intelligence  \textbf{34}(07),  12805--12812 (Apr
  2020)

\bibitem{passport2}
Zhang, J., Chen, D., Liao, J., Zhang, W., Hua, G., Yu, N.: Passport-aware
  normalization for deep model protection. In: Advances in Neural Information
  Processing Systems (NeurIPS) (2020)

\bibitem{secret_reveal}
Zhang, Y., Jia, R., Pei, H., Wang, W., Li, B., Song, D.: The secret revealer:
  Generative model-inversion attacks against deep neural networks. In:
  Proceedings of the IEEE/CVF Conference on Computer Vision and Pattern
  Recognition (CVPR) (June 2020)

\end{thebibliography}

\end{document}